# Argos: A Decentralized Federated System for Detection of Traffic Signs in CAVs

Seyed Mahdi Haji Seyed Hossein
ECE Department
University of Tehran
Tehran, Iran
mehdihaji@ut.ac.ir

Alireza Hosseini
ECE Department
University of Tehran
Tehran, Iran
alyreza.hosseini@ut.ac.ir

Soheil Hajian Manesh
ECE Department
University of Tehran
Tehran, Iran
soheilhajian@ut.ac.ir

Amirali Shahriary
ECE Department
University of Tehran
Tehran, Iran
amirali.shahriary@ut.ac.ir

*Abstract*—Connected and automated vehicles generate vast amounts of sensor data daily, raising significant privacy and communication challenges for centralized machine learning approaches in perception tasks. This study presents a decentralized, federated learning framework tailored for traffic sign detection in vehicular networks to enable collaborative model training without sharing raw data. The framework partitioned traffic sign classes across vehicles for specialized local training using lightweight object detectors, aggregated model parameters via algorithms like FedProx, FedAdam and FedAVG in a simulated environment with the Flower framework, and evaluated multiple configurations including varying server rounds, local epochs, client participation fractions, and data distributions. Experiments demonstrated that increasing server rounds from 2 to 20 boosted accuracy from below 0.1 to over 0.8, moderate local epochs (8–10) provided optimal efficiency with accuracies around 0.67, higher client participation fractions enhanced generalization up to 0.83, FedProx outperformed other aggregators in handling heterogeneity, non-IID data distributions reduced performance compared to IID, and training duration primarily scaled with the number of rounds rather than aggregation strategy. We conclude that this federated approach may offer a scalable, privacy-preserving solution for real-world vehicular deployments, potentially guiding future integrations of robust aggregation and communication optimizations to advance intelligent transportation systems.

## I. Introduction

Connected and automated vehicles (CAVs) generate enormous volumes of sensor data on the order of TB per vehicle per day to support perception, navigation, and safety functions. This data often includes sensitive information (e.g. location, driver behavior) and is subject to strict privacy regulations (such as GDPR in Europe)[1]. Traditional machine learning (ML) approaches rely on aggregating all raw vehicle data on a central server for model training, but this raises serious privacy and security concerns. Federated learning (FL) has emerged as a solution: each vehicle trains a local model on its own data and only shares model updates with a server, so that raw images or telemetry never leave the vehicle[2]. By keeping image data (for example from onboard cameras) locally on each vehicle and exchanging only model weights, FL can protect user privacy while greatly reducing the communication burden of transmitting high-bandwidth data[3].

FL and related distributed methods have been widely explored in the Internet of Vehicles (IoV) domain. For example, Quéméneur *et al.* note that connected vehicles can leverage abundant edge data for ML, and that FL is a "promising solution to train sophisticated ML models in vehicular networks while protecting the privacy of road users and mitigating communication overhead". Numerous studies have applied FL to vision-based perception tasks in vehicles: for instance, road-damage and pothole detection, semantic segmentation, and general object detection have all been demonstrated with a federated framework. These works show that collaborative learning across many vehicles can improve model accuracy and robustness without requiring raw image sharing[3]. FL-based systems have also been proposed for traffic management and prediction problems (e.g. demand forecasting and congestion control), highlighting the flexibility of FL for IoV applications. In each case, local model training on vehicles allows leveraging diverse driving data while keeping user data on board[4].

Despite these advances, most prior FL approaches make assumptions that are unrealistic in real vehicular settings. In particular, many methods still assume a central coordinator or fully shared label space. For example, several works require that every vehicle send its annotations to a central server for labeling or that each client can train on all object classes (e.g. all traffic sign types)[2]. In practice, each vehicle only observes a subset of sign classes in its local environment, so requiring full-class data on each vehicle is impractical. Moreover, many FL studies presume ideal communication (synchronous updates, high bandwidth links) when in fact vehicle-to-infrastructure networks have limited capacity and frequent disconnects. These gaps in the reliance on centralized labeling, full-class training assumptions, and unlimited communication motivate the need for a truly decentralized FL framework that adapts to the constraints of CAVs.

In this paper, we propose a **decentralized federated learning framework** for traffic sign annotation and detection in connected and automated vehicles. We explore multiple experimental configurations to evaluate the framework under realistic IoV conditions, which are detailed in the *Methodology* section. In our approach, vehicles exchange updates for their specialized classifiers via vehicle-to-infrastructure communication links, without sharing raw sensor data. Our main contributions are threefold: (1) the design of **a fully decentralized FL scheme tailored to traffic sign detection in CAV environments;** (2) **a performance evaluation showing that this approach matches** or surpasses centralized training accuracy while reducing communication overhead and preserving privacy; and (3) **a comparative analysis of different operating configurations for CAVs** to determine the most effective strategy for real-world deployment.

The remainder of this paper is organized as follows. Section II reviews related work on federated learning for connected vehicles and visual perception tasks. Section III outlines the methodology and describes our proposed federated framework in detail. Section IV presents the results evaluating detection accuracy and communication efficiency. Finally, Section V concludes the paper by summarizing key insights and suggesting avenues for future research.

## II. LITERATURE REVIEW

Federated learning (FL) has recently emerged as a promising paradigm for training machine learning models across distributed vehicles while preserving data privacy and distributing computation across all contributors. In connected and automated vehicles (CAV), onboard sensors (cameras, LiDAR, etc.) generate vast amounts of data for perception, planning, and control. Centralized training of such data is problematic due to privacy concerns (e.g. sensitive location or personal data), communication overhead, and data silos. Surveys by Chellapandi *et al.* note that FL allows multiple vehicles to collaboratively train shared models, thereby leveraging diverse driving environments and improving overall performance while securing local data. This aligns with analyses in intelligent transportation systems (ITS), which highlight that traditional centralized approaches suffer from poor real-time performance, data fragmentation, and privacy risks [1]. By contrast, FL's decentralized training, where each vehicle (or edge client) updates a local model and only shares weight updates, directly addresses these issues. The foundational FedAvg algorithm by McMahan *et al.* exemplifies this: it iteratively averages local model updates across clients and has been shown robust to non-IID data, though Non-IID distributions may require additional communication rounds, while drastically reducing required communication compared to naïve SGD[5]. Federated frameworks in vehicles can be centralized (with a single aggregator) or decentralized (e.g. gossip protocols). Rjoub *et al.* compare these approaches: in adverse-weather detection, they show a federated approach outperforms both a centralized model and a fully decentralized gossip scheme[6]. Together, these works establish FL as an enabling technology for CAVs, trading some training efficiency for privacy and scalability.

### A. FL for Object and Scene Perception

A major application of FL in vehicles is real-time object detection. Traditional detectors (e.g. R-CNN variants) were accurate but relatively slow, whereas the YOLO family ("You Only Look Once") is designed for real-time, high-speed detection and is now widely used in autonomous vehicles[4]. Researchers have begun federating these modern detectors. For example, Rjoub *et al.* propose a Federated YOLO system to improve object detection in heavy snow. Using the YOLO CNN at each vehicle with a federated aggregation step, they show improved detection accuracy in snow compared to a fully centralized model or a gossip-based scheme[6]. More recently, Quéméneur and Cherkaoui introduced FedPylot, the first large-scale federated YOLOv7 framework for Internet-of-Vehicles (IoV) object detection. FedPylot specifically tackles *data heterogeneity,* including unbalanced client data, concept drift, and label skew, which are inherent in vehicles from different locations and times. It uses an MPI-based simulator to emulate federated clients and implements hybrid encryption for secure updates. Their study evaluates not only accuracy but also communication cost and inference speed, demonstrating a balanced trade-off for autonomous vehicle applications[3]. These works underscore that FL can extend state-of-the-art object detectors (like YOLO) across a fleet of vehicles, enabling privacy-preserving joint training of perception models[6].

Federated schemes have also been applied to other perception tasks. Traffic sign recognition is critical for safety. Xie *et al.* propose an FL framework using spiking neural networks (SNNs) instead of conventional CNNs. SNNs are energy-efficient, making them attractive for vehicles with limited power. In their system, each vehicle encodes traffic sign images into spike-based inputs and trains an SNN locally. The federated SNN achieves higher accuracy and better noise immunity than a federated CNN baseline, while consuming less energy[7]. In the in-cabin domain, deep learning models monitor driver and passenger behavior. For instance, Kim *et al.* develop a 3D human pose estimation and seat-belt segmentation network for real-time driver monitoring[8]. Although this work itself is not federated, it highlights the range of vision tasks in vehicles (and the potential to federate similar models) More directly, Lindskog *et al.* apply FL to driver drowsiness detection using the YawDD dataset. By federating CNN-based drowsiness models across cars, they achieve ~99.2% accuracy on detecting driver fatigue – comparable to centralized training – while keeping data local[9].

Finally, FL has been explored in environmental awareness and V2V applications. Lin and Liang address the "vehicle identification" problem: linking V2V message senders to the corresponding vehicle in an image. They train a supervised identification model using both FL and automatic labeling: vehicles keep their image data private, and an opportunistic auto-labeling scheme annotates which vehicles are sending signals. This hybrid approach overcomes drivers' reluctance to share images or labels, while still leveraging FL's distributed training[10]. Such work demonstrates how FL can integrate vision data with inter-vehicle communications to enhance situational awareness without compromising privacy.

### B. Communication and Privacy Considerations

The literature consistently notes that communication efficiency is a key challenge. Vehicles may have intermittent connectivity and limited bandwidth, making it critical to reduce the number of FL rounds or the size of model updates. FedAvg itself reduces communication by 10–100× compared to standard SGD[5], but researchers continue to investigate compression, asynchronous updates, and selective client participation to further cut overhead. In the automotive context, FedPylot explicitly measures communication cost, and Rjoub *et al.* compare federated vs. centralized vs. gossip communication schemes[3][6] Fault tolerance and heterogeneous client capabilities are also active concerns (as surveyed by Zhang *et al.*[11] and Chellapandi *et al.*), since a vehicle joining late or dropping out should not derail training [1].

Privacy and security are equally important. FL inherently avoids raw data sharing, but model updates can still leak information. To mitigate this, FedPylot incorporates a hybrid encryption layer for server–client messages[3]. Other works in FL for object detection propose differential privacy: for instance, Wang *et al.* introduce a federated object detector that adaptively injects noise during training to obscure individual data, balancing privacy with accuracy[12]. Surveys recommend additional safeguards (blockchain, secure aggregation, anomaly detection) to detect malicious clients or ensure confidentiality[10]. Overall, the literature stresses that FL in vehicles must be communication-efficient and privacy-preserving, leveraging techniques like encryption and differential privacy while still achieving high perception performance.

### III. METHODOLOGY

In this section, we present the architecture, data processing pipeline, and training strategy for our decentralized federated learning software architecture for traffic sign detection in connected vehicles.

#### A. Dataset and Preprocessing

We utilize the **Traffic Signs Dataset - Mapillary and DFG** [1], which contains over 19,000 high-resolution road images and more than 30,000 annotated traffic signs across 76 distinct classes.

Each annotation is provided in YOLO format with normalized bounding box coordinates [xcenter,ycenter,width,height] for each bounding box and corresponding class indices. The dataset structure includes cropped traffic signs, full scene images, YOLO .txt files for bounding boxes, and class metadata.

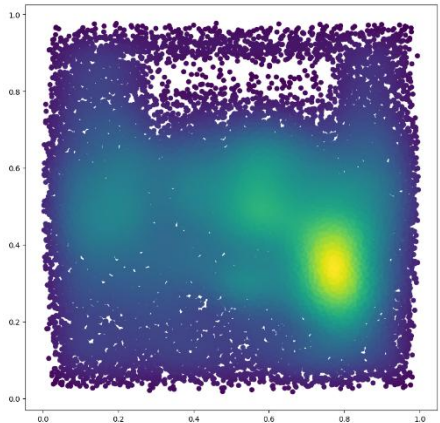

*Figure 1 - Distribution of data points in the (x_center, y_center) plane*

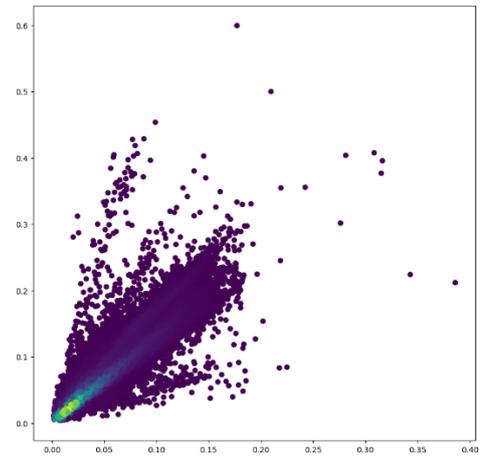

*Figure 2 - Scatter plot of width against height*

To facilitate detection tasks, we first parse YOLO label files and generate a unified dataframe containing image paths, bounding box coordinates, class labels, and annotation counts. Exploratory data analysis is performed to assess class distribution, bounding box dimensions, and object spatial density. Data visualization includes KDE plots for center and dimension densities, and bounding box rendering.

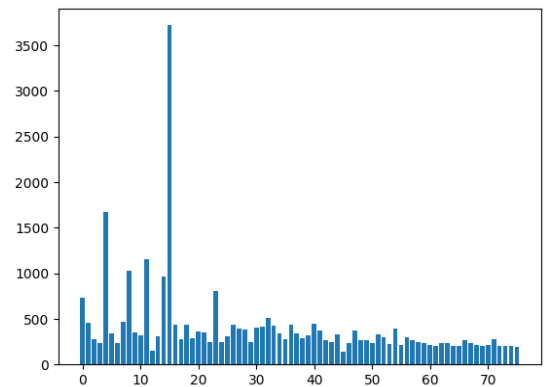

*Figure 3 - Number of each classes*

---

[1] https://www.kaggle.com/datasets/nomihsa965/traffic-signs-dataset-mapillary-and-dfg

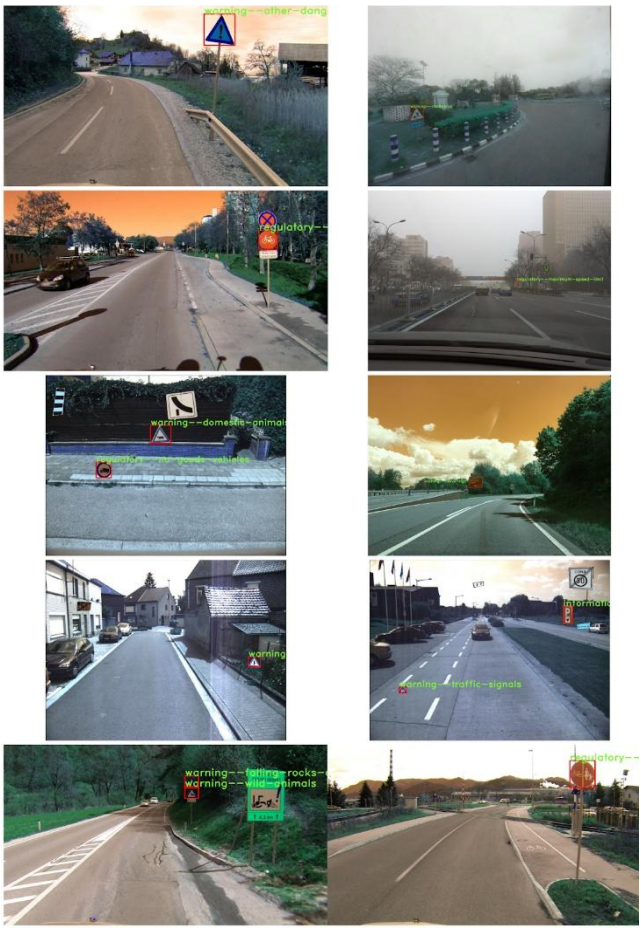

*Figure 4 - Some samples, note: there may be some signs on the road or somewhere you may not expect, this is because of data augmentation in this dataset.*

### B. Task Distribution and Annotation Design

We explored several approaches for distributing tasks among participants. In the task distribution process, we examined the number of server rounds. In each round, the server selects a fraction of clients to participate in the training step. The number of rounds indicates how many times this process—selecting a fraction of clients and training the model with those participants—is repeated.

Another factor we examined was the distribution algorithm. We considered different data distributions, such as Non-IID and IID [5]. Non-IID indicates that the data is not identically distributed across clients, whereas IID implies that the data samples on each client are independent and identically distributed.

### C. Local Model Training

Each vehicle trains an instance of a lightweight object detection model (e.g., fasterrcnn_mobilenet_v3_large_320 or similar architectures) on its locally collected and labeled dataset. Given that the partitioned data are smaller and computationally more feasible for deployment on embedded edge devices. Data augmentation techniques, such as scaling, flipping, and blurring, are applied to compensate for limited sample sizes in narrow-class local datasets.

Training is performed independently on each vehicle using standard object detection loss functions, with early stopping based on validation performance. During this stage, we also explored different settings, including the number of local epochs for each vehicle and the fraction of participating clients in each server round. This accounts for real-world variability, such as some drivers being unavailable during a particular server round.

### D. Federated Aggregation

Once local training is complete, each vehicle transmits only model parameters (weights) — not raw images or labels — to a central aggregator server. Another factor we examined was the aggregation algorithm. We tested several approaches, including the standard Federated Averaging (FedAvg)[5], FedAdam [13], and FedProx [14]. The aggregator then computes a weighted aggregation of the parameters according to the selected algorithm to update the global model, which is subsequently shared back with participating vehicles for the next training round. For this purpose, we used **Flower Framework**[2].

After aggregation, the updated global model is redistributed to all participating vehicles for the next training epoch, creating an iterative improvement cycle. This continues for a predefined number of federated rounds or until convergence.

### E. Experimental Setup

In this section, we present the results of our experiments under multiple configurations, with the goal of identifying the most effective setup for real-world deployment. Across all experiments, the learning rate was fixed at 0.001, with a batch size of 4, and the evaluation fraction was set to 1.0, meaning all clients participated in the evaluation stage. Training was performed using the Adam optimizer implemented in PyTorch. For federated learning, we employed the widely used Flower framework, simulating the environment on a single NVIDIA Tesla P100 GPU. The GPU resources were partitioned proportionally to the number of clients, and each client was assigned a dedicated CPU to ensure fairness and consistency in the simulation. Also, at the evaluation stage, we used 0.1 IoU for calculating accuracy.

## IV. RESULT

**Figure 5** illustrates the impact of varying the number of server rounds on model accuracy when using the FedAdam aggregation algorithm. As expected, models trained with a greater number of server rounds achieve higher accuracy. With only 2 rounds, accuracy remains very low (<0.1), indicating insufficient global coordination. Increasing the number of rounds to 5 results in a sharp improvement, with accuracy stabilizing around 0.55. When extended to 10 and 20 rounds, the model continues to improve steadily, ultimately reaching accuracies above 0.8. These results highlight the critical role of server rounds in federated learning: more global communication cycles allow the system to better align local models, reduce divergence caused by IID data, and converge toward an accurate global model.

---

[2] flower.ai

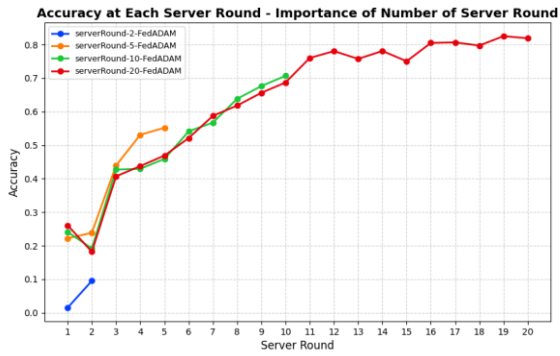

*Figure 5 - Impact of the number of server rounds on model accuracy using the FedAdam aggregation algorithm. All other parameters — including aggregation method, number of local epochs, and client participation fraction — were held constant to isolate the effect of server rounds.*

The effect of varying the number of local epochs on model accuracy is shown in **Figure 6**. Accuracy consistently improves as the number of local training epochs increases. At just one epoch per client, the model achieves ~38% accuracy by the 8th server round, while increasing to 10 epochs raises performance to ~67%. Notably, the improvement from 10 to 20 epochs is marginal (66.7% vs. 67.0%), suggesting diminishing returns beyond a certain point. This indicates that while additional local training strengthens client contributions, excessive epochs do not proportionally enhance global performance and may unnecessarily increase computation time on client devices. Thus, an intermediate choice (e.g., 8–10 epochs) appears to provide the best trade-off between accuracy and computational efficiency.

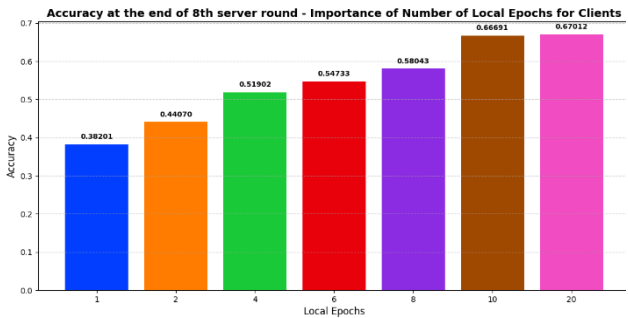

*Figure 6 - Impact of the number of local training epochs per client on model accuracy, measured at the end of the 8th server round. All other experimental factors — including aggregation algorithm, client participation fraction, and number of server rounds — were held constant. Only the number of local epochs was varied to isolate its effect.*

The effect of varying the client participation fraction on model accuracy is shown in **Figure 7**. Accuracy improves significantly as more clients are selected for training in each round. With only 10% of clients participating (2 out of 20), the accuracy reaches ~55% by the 8th server round. Increasing the fraction to 50% raises accuracy to ~62%, while full client participation (100%) achieves the highest accuracy of ~83%. This trend highlights the importance of involving a larger portion of clients in each round, as more diverse data contributions improve generalization. However, higher participation also increases communication and computational overhead, suggesting that intermediate fractions (e.g., 70–90%) may provide a practical balance between performance and efficiency.

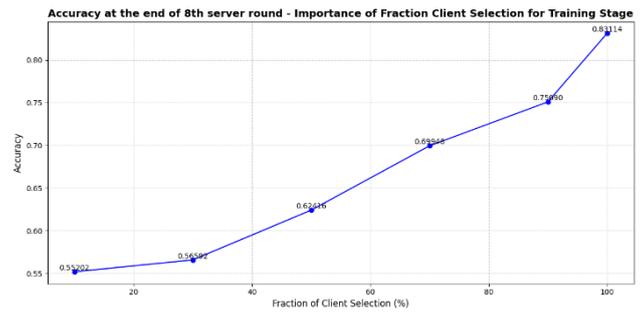

*Figure 7 - Impact of client participation fraction on model accuracy, measured at the end of the 8th server round. The total number of clients in the experiment was 20, and only the participation fraction was varied, while all other factors (aggregation algorithm, local epochs, and number of server rounds) were kept constant.*

The investigation into the influence of various server aggregation strategies on the performance of a federated learning model reveals distinct trends in accuracy across server rounds. FedAVG, despite its common use, showed an erratic and ultimately disappointing performance, with accuracy fluctuating significantly before settling at a low point by the final round. This suggests it may not be a suitable optimizer for the complex dynamics of this traffic sign detection task. In contrast, the FedProx strategy displayed a more encouraging trajectory, starting with a respectable accuracy and demonstrating a strong, positive growth trend, culminating in the highest accuracy of all strategies tested. This indicates its effectiveness in handling client data heterogeneity, which is a common challenge in federated learning. Meanwhile, the FedADAM approach presented a more stable and consistently promising trend. Although not reaching the peak accuracy of FedProx, its steady and reliable climb throughout the server rounds makes it a reliable and optimistic choice for scenarios where consistent performance gains are prioritized over peak, potentially unstable, results.

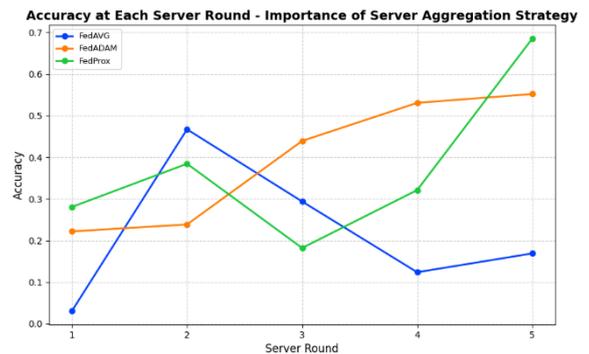

*Figure 8 - Accuracy at Each Server Round – Importance of Server Aggregation Strategy. This graph illustrates the accuracy of the federated learning model for traffic sign detection over five server rounds, comparing three different server aggregation strategies: FedAVG, FedADAM, and FedProx. All other experimental factors, including client selection, local epochs, and learning rates, were kept constant to isolate the impact of the aggregation strategy. Accuracy was measured at the end of each server round, with the final results shown at the end of the 5th server round.*

The comparative analysis between **Independent and Identically Distributed (IID)** and **Non-Independent and Identically Distributed (Non-IID)** data distributions highlights a significant challenge in federated learning. The model trained on the IID dataset demonstrates a consistent and superior performance, with its accuracy curve showing a steady upward trend. This is expected, as each client's data is a miniature, representative sample of the overall dataset, leading to more stable and aligned local model updates. In stark contrast, the model trained on the Non-IID dataset exhibits a less favorable performance. Its accuracy remains notably lower than the IID model and shows a more volatile trajectory, indicating that the statistical heterogeneity across clients' local datasets negatively impacts the global model's ability to generalize. This is a common issue in practical federated learning scenarios, as client data (e.g., from different vehicles in different regions) is inherently skewed. The substantial performance gap between the two demonstrates the need for advanced aggregation strategies and fairness-aware techniques to mitigate the negative effects of non-IID data.

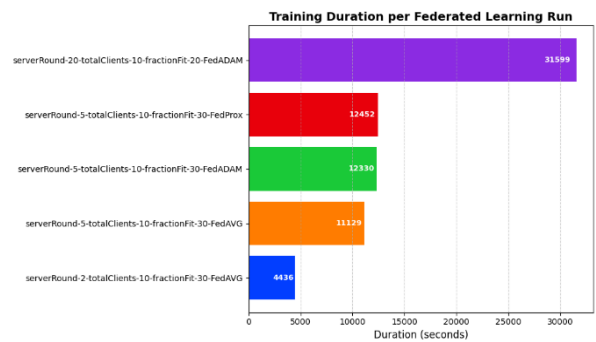

*Figure 10 - Training Duration of Different Federated Learning Runs. This bar chart illustrates the total training duration in seconds for several federated learning experiments, each with a unique combination of server rounds and aggregation strategies. The duration is measured from the start of the first server round to the completion of the final round. The figure highlights how varying key parameters, such as the number of server rounds and the chosen aggregation strategy, impacts the overall training time for the federated learning model.*

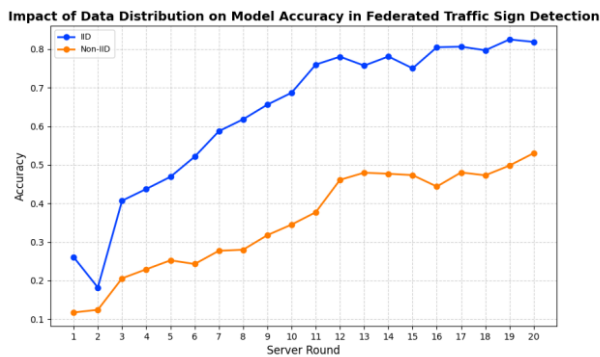

*Figure 9 - Performance Comparison of IID vs. Non-IID Data Distribution. This figure presents a comparative analysis of model accuracy over eight server rounds for traffic sign detection, contrasting IID (Independent and Identically Distributed) and Non-IID (Non-Independent and Identically Distributed) data distributions. All other experimental parameters, such as the aggregation strategy, number of clients, and local epochs, were kept constant. The observed accuracy was measured at the end of the 8th server round.*

The analysis of training duration across various federated learning runs reveals that the number of server rounds is the primary determinant of the overall training time. As seen in the figure, a run with 20 server rounds took over 31,000 seconds, significantly longer than the runs with just 5 server rounds, which averaged around 12,000 seconds. Interestingly, for a fixed number of server rounds (e.g., the three 5-round runs), the choice of aggregation strategy—FedAVG, FedADAM, or FedProx—had a minimal impact on the total duration. Their training times were very similar, suggesting that the computational overhead introduced by these specific algorithms is negligible compared to the total time spent across all communication rounds and local training stages. This finding is crucial for resource planning in real-world deployments, as it confirms that the number of global communication rounds, rather than the specific aggregation method, is the key factor in estimating training costs.

## V. CONCLUSION

This paper investigated a decentralized federated learning framework for traffic-sign detection in connected and automated vehicles. The primary goal was to determine which FL settings outperform —where each vehicle labels and trains only on the sign classes it observes, and only model parameters are exchanged—could produce competitive global detectors while preserving privacy and reducing communication.

Our experiments showed that global coordination was the dominant factor for success: increasing the number of server rounds substantially improved accuracy, whereas too few rounds prevented meaningful learning. Moderately increased local computation (≈8–10 epochs) yielded most of the benefit, with diminishing returns beyond that point. Broader client participation improved generalization, and aggregation schemes that account for client heterogeneity (e.g., FedProx) outperformed vanilla FedAvg in this non-IID, class-partitioned setting; FedAdam produced steady, stable gains. Non-IID data distributions degraded convergence and final accuracy, highlighting the well-known heterogeneity challenge in FL. Training time was driven mainly by the number of global rounds rather than the specific aggregation algorithm. Together, these findings support the viability of class-specialized FL provided sufficient rounds, representative client sampling, and heterogeneity-aware aggregation.

A practical limitation is that our evaluation used a simulated Flower environment on a single GPU and a relaxed IoU threshold, so real-world networking, latency, and standardized detection metrics need further validation.

Implicitly, the results suggest concrete deployment guidelines: favor moderate local work, ensure diverse client participation, and adopt aggregation methods robust to skew. Future work should validate the approach on real vehicular testbeds, report standard detection metrics (mAP at IoU 0.5/0.75), integrate stronger privacy mechanisms (secure aggregation / differential privacy), explore communication reduction (sparsification, quantization, distillation), and evaluate energy-efficient on-vehicle models (e.g., SNNs) for sustainable deployment.

ACKNOWLEDGMENT

We gratefully acknowledge Dr. Mehdi Modarressi for his guidance and support throughout this research.